\def\year{2020}\relax
\documentclass[letterpaper]{article} 
\usepackage{aaai20}  
\usepackage{times}  
\usepackage{helvet} 
\usepackage{courier}  
\usepackage[hyphens]{url}  
\usepackage{graphicx} 
\usepackage{subcaption}
\usepackage{graphicx}  
\title{PCGRL: Procedural Content Generation via Reinforcement Learning}

\author{Ahmed Khalifa, Philip Bontrager, Sam Earle, Julian Togelius\\ 
New York University\\
ahmed@akhalifa.com, philipjb@nyu.edu, smearle93@gmail.com, julian@togelius.com
}

\begin{document}

\maketitle

\begin{abstract}
We investigate how reinforcement learning can be used to train level-designing agents. This represents a new approach to procedural content generation in games, where level design is framed as a game, and the content generator itself is learned. By seeing the design problem as a sequential task, we can use reinforcement learning to learn how to take the next action so that the expected final level quality is maximized. This approach can be used when few or no examples exist to train from, and the trained generator is very fast. We investigate three different ways of transforming two-dimensional level design problems into Markov decision processes, and apply these to three game environments.
\end{abstract}

\section{Introduction}
Reinforcement learning is commonly used to learn to play games, which makes sense as the problem of playing a game can easily be cast as a reinforcement learning problem; the action space is simply the actions available to the agent, and most games have a score or similar which can be used to provide a reward signal~\cite{justesen2019deep}. In contrast, problems of designing games or game content are most often cast as optimization processes, where a measure of quality is used as an objective function~\cite{togelius2011search}, or sometimes as supervised learning problems~\cite{summerville2018procedural}. In the game industry, where many games rely on content generation, this process is typically performed by hand-crafted heuristic algorithms~\cite{shaker2016procedural}.

In this paper, we investigate how to generate game levels using reinforcement learning. To the best of our knowledge, this is the first time reinforcement learning is brought to bear on this problem. This is probably because it is not immediately obvious how to cast a level generation problem as a reinforcement learning problem. The core question that this paper attempts to answer is how level generation can be formulated as a tractable reinforcement learning problem. We formulate observation spaces, action spaces and reward schemes so as to make existing RL algorithms learn policies that result in high-quality game levels.

Conceptually, the main difference to existing approaches to Procedural Content Generation (PCG) is that we do not search the space of game content, but rather the space of policies that generate game content. At each step, the policy is asked to take the action that leads to the highest expected final level quality. This can be contrasted to search-based approaches where each ``action'' generates a complete level, or to approaches based on supervised or unsupervised learning where complete levels are sampled from a learned model.

Reinforcement learning approaches to PCG have several potential advantages over existing methods. Compared to search-based methods~\cite{togelius2011search}, machine learning approaches can generate new levels, after training, much faster as no search is needed on demand. This comes at a cost of having a long training phase which search-based PCG methods do not have, so we move the time consumption from inference to training. Compared to supervised learning ~\cite{summerville2018procedural}, the big advantage is that no training data is necessary. Another advantage is that the incremental nature of the trained policies makes them potentially more suitable to interactive and mixed-initiative approaches to PCG, where content is created together with human users~\cite{yannakakis2014mixed}.

In our experiments, we focus on two-dimensional levels for three different game environments. Two of these are actual games -- the classic puzzle game Sokoban (Thinking Rabbit, 1982) and a simplified version of the Legend of Zelda (Nintendo, 1986) -- and the third is a simple maze environment where the objective is to generate mazes containing long paths. We formulate three different representations of game levels as reinforcement learning problems: the narrow representation, the turtle representation, and the wide representation (which we will discuss later in the paper).
We find that all three representations can be successful on all three game scenarios, given that the right choices are made regarding reward schemes and episode lengths, but that there are interesting differences in the generated artefacts.

\section{Background}

Procedural level generation research has started to incorporate more machine learning techniques~\cite{jain2016autoencoders,volz2018evolving} after the recent advances in machine learning. Reinforcement Learning (RL), though, was rarely applied to PCG, even after its success in playing video games~\cite{mnih2015human},
perhaps because it is unclear how to form the level generation process as an RL problem.


One solution is to frame the content generation as an iterative process, where at each step the agent is trying to modify a small part of the content, similar to the idea from \citeauthor{guzdial2018co}'s~\shortcite{guzdial2018co} work. Several researchers have explored the idea of iterative generation to build a platformer level using supervised sequence learning methods, such as Markov Chains~\cite{snodgrass2014experiments} and 
LSTMs~\cite{summerville2016super}. 

Looking at level generation from a sequential perspective makes it possible to formulate it as a Markov Decision Process (MDP) where the agent is making small, iterative changes to improve the current level. For example, \citeauthor{mcdonald2019markov}~\shortcite{mcdonald2019markov} formulates the process of 3D building-generation as a MDP. They define the state space as the locations of several cuboids, while the action space corresponds to movement of these cuboids along any of the Cartesian axes. They reward actions that minimize intersection between the cuboids while making them touch. They don't do any learning; instead they use a greedy agent that picks the next action to maximize the immediate reward.

\citeauthor{earle2019using}~\shortcite{earle2019using} trains fractal neural networks using A2C~\cite{mnih2016asynchronous} to play SimCity (Will Wright, 1989). Although this work doesn't tackle the idea of using RL for PCG, we can look at it as a step toward PCGRL, as the trained models design cities, which could constitute game levels. The only work we found that used RL to directly generate game content was that of \citeauthor{chen2018q}~\shortcite{chen2018q} and \citeauthor{guzdial2019friend}~\shortcite{guzdial2019friend}. In \citeauthor{chen2018q}'s work, Q-Learning~\cite{watkins1992q} is used to train a deck-building system that outperforms search-based methods.
In \citeauthor{guzdial2019friend}'s work, they proposed a mixed initiative tool to design levels for Super Mario Bros (Nintendo, 1985) that uses active learning to update the trained models to adapt to the user choices. Although the paper didn't use RL to train a model from scratch, 
the system was successfully adapting to the users choices. 


\section{PCGRL Framework}

\begin{figure}
    \centering
    \includegraphics[width=0.7\linewidth]{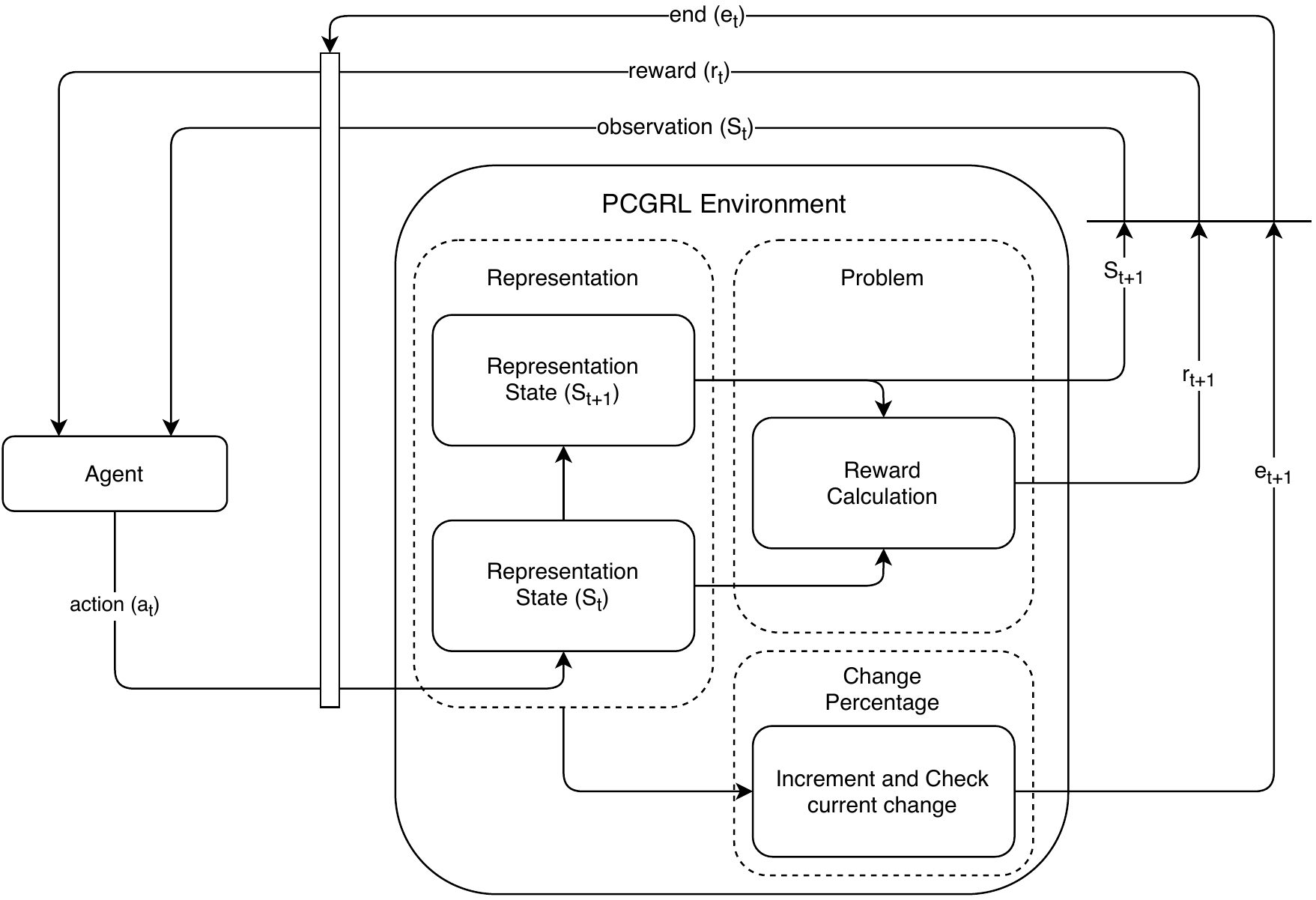}
    \caption{The system architecture for the PCGRL environment for content generation.}
    \label{fig:pcgrl}
\end{figure}

The PCGRL Framework casts the PCG process as an iterative task 
instead of generating the whole content at once. 
We thus see content generation as a MDP, where at each step the agent gets an observation and reward then responds with an action. In this work, we will only be looking at the task of level generation, but everything we are going to discuss here can be applied to other types of content generation.

To realize the idea of iterative content generation, we start with a level populated by random tiles. At each step, the agent is allowed to make a small change in the level (such as one tile). This change will be judged by the system with respect to a target goal for the level, and assigned a reward. The reward should reflect how much closer that small change has brought the agent to its goal state. For example: if we are generating a PacMan (NAMCO, 1980) level, one of the goals is to have only one player; so a change that adds a player when there is none is a positive change, and negative otherwise. The system should also determine the halting point of the generation process (limiting the number of iterations so that it doesn't take forever).

To make the framework easy to implement for any game, we break it down into three parts, and isolate game-related information from the generation process. These parts are: the Problem module, the Representation module, and the Change Percentage. The problem module stores information about the generated level such as goal, reward function, etc. The representation module is responsible for transforming the current level into a viable observation state for the agent. The change percentage limits the number of changes the agent can affect in the content over the course of an episode, preventing it from changing the content too much. 

Figure~\ref{fig:pcgrl} shows the PCGRL agent-environment loop. The agent observes the current state ($S_{t}$), and based on it, sends an action ($a_{t}$) to the representation module, which in turn transforms the state ($S_{t}$) into the next state ($S_{t+1}$). These two states ($S_{t}$ and $S_{t+1}$) are both sent to the problem module, which assesses the change's effect on map quality and returns the new reward ($r_{t+1}$). This new reward and the new state ($S_{t+1}$) back to the agent, and the loop continues. The loop could terminate early by sending an end signal ($e_{t+1}$) if it makes a lot of changes to the map.

After training, these agents are used as generators: they iterate over a randomly-initialized map for a fixed number of steps; improving it slightly or transforming it completely.

\subsection{Problem}
The problem module is responsible for providing all the information about the current generation task. For example: if we were trying to generate a Super Mario Bros (Nintendo, 1985) level, the problem module would support us with the level size, the types of objects that can appear in the level, etc. This module provides two functions. The first function assesses the change in quality of generated content after a certain agent action. For example: if the agent removes an object from a game level, the problem module will assess the resultant change in level quality and return a reward value that the agent can use to learn. 

The second function determines when the goal is reached, which terminates the generation process. For example: if we are generating a house layout, we can terminate generation after we have a certain number of rooms created. It is important to define a goal for our problem that leads to many possible levels, as we are trying to learn a generator and not find a single level. For example: if the goal function for a 10x10 maze is to have a level with path length equal to 54 (which is the longest possible path between any two tiles on such a map), then the model will learn to generate only these maps. This is fine if the designer wants to find a single level, but there are many optimization techniques for that and here we focus on learning to generate multiple distinct levels.

\subsection{Representation}

\begin{figure}
    \centering
    \includegraphics[width=0.6\linewidth]{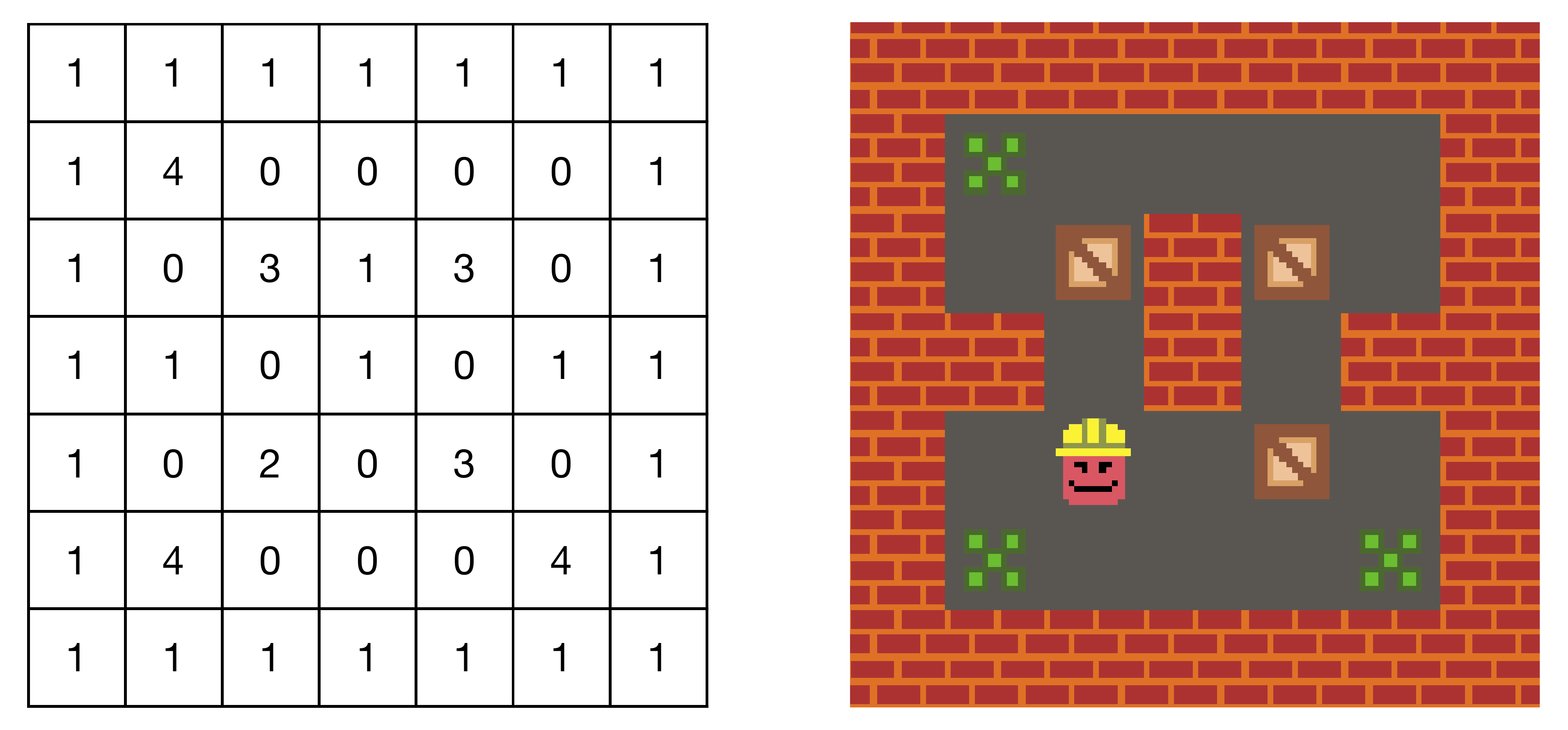}
    \caption{Sokoban level as 2D integer array.}
    \label{fig:sokoban_array}
\end{figure}

To model content generation as a MDP, we need to define the state space, action space, and transition function. The representation module is responsible for this transformation. Its role is to initialize the problem, maintain the current state, and modify the state based on the agent's action.

For the sake of simplicity, we represent a generated level as a 2D array of integers where the location in the array corresponds to the location in the level, and the value defines the type of object in that location. For example: Figure~\ref{fig:sokoban_array} shows a Sokoban (Thinking Rabbit, 1982) level as 2D array, and the corresponding level. This constraint makes it easy for us to adopt the same representations that were used in the work by~\citeauthor{bhaumik2019tree}~\shortcite{bhaumik2019tree} on generating levels using tree search algorithms. In that work, they defined the following representations:
\begin{itemize}
    \item \textbf{Narrow:} the simplest way of representing the problem. It is inspired by cellular automata~\cite{wolfram1983cellular} where at each step, the agent is given the current state and a location. It then is allowed to make a change at that location. The observation space is defined as the current state (as the 2D integer array) and the modification location (as an x and y index in the 2D array). The action space is defined as a no-action (which skips the current location) or a change tile action (value between $0$ to $n-1$ where $n$ is the number of tile types provided by the problem module). 
    \item \textbf{Turtle:} inspired from the turtle graphics languages such as Logo~\cite{bolt1967logo} where at each step, the agent can move around and modify certain tiles along the way. The observation space is represented as the current state (as the 2D integer array) and the current agent location (as an x and y index in the 2D array). The action space is defined as: a movement action (which moves the agent either $up$, $down$, $left$, or $right$) or a change tile action (value between $0$ to $n-1$ where $n$ is the number of different objects provided by the problem module).
    \item \textbf{Wide:} is similar to \citeauthor{earle2019using}'s work~\shortcite{earle2019using} on playing SimCity. At each step, the agent has full control over the location and tile type. The observation space is the current state (as the 2D integer array). The action space is defined as the affected location ($x$ and $y$ position on the level) and the change tile action (value from $0$ to $n-1$ where $n$ is the number of tile types provided by the problem module).
\end{itemize}

Each representation corresponds to a distinct class of agents. Those with \textbf{Narrow} representations are beholden to a predetermined sequence of build-locations; those with \textbf{Turtle} representations have local control over the current location, but only relative to the last; and those with \textbf{Wide} representations have full control. We could also develop hybrid representations (e.g. a mix of narrow and wide, where the agent can modify a small area around a given location) or modify their action schemes (e.g. changing multiple tiles instead of a single tile).

\begin{figure}
    \centering
    \includegraphics[width=0.6\linewidth]{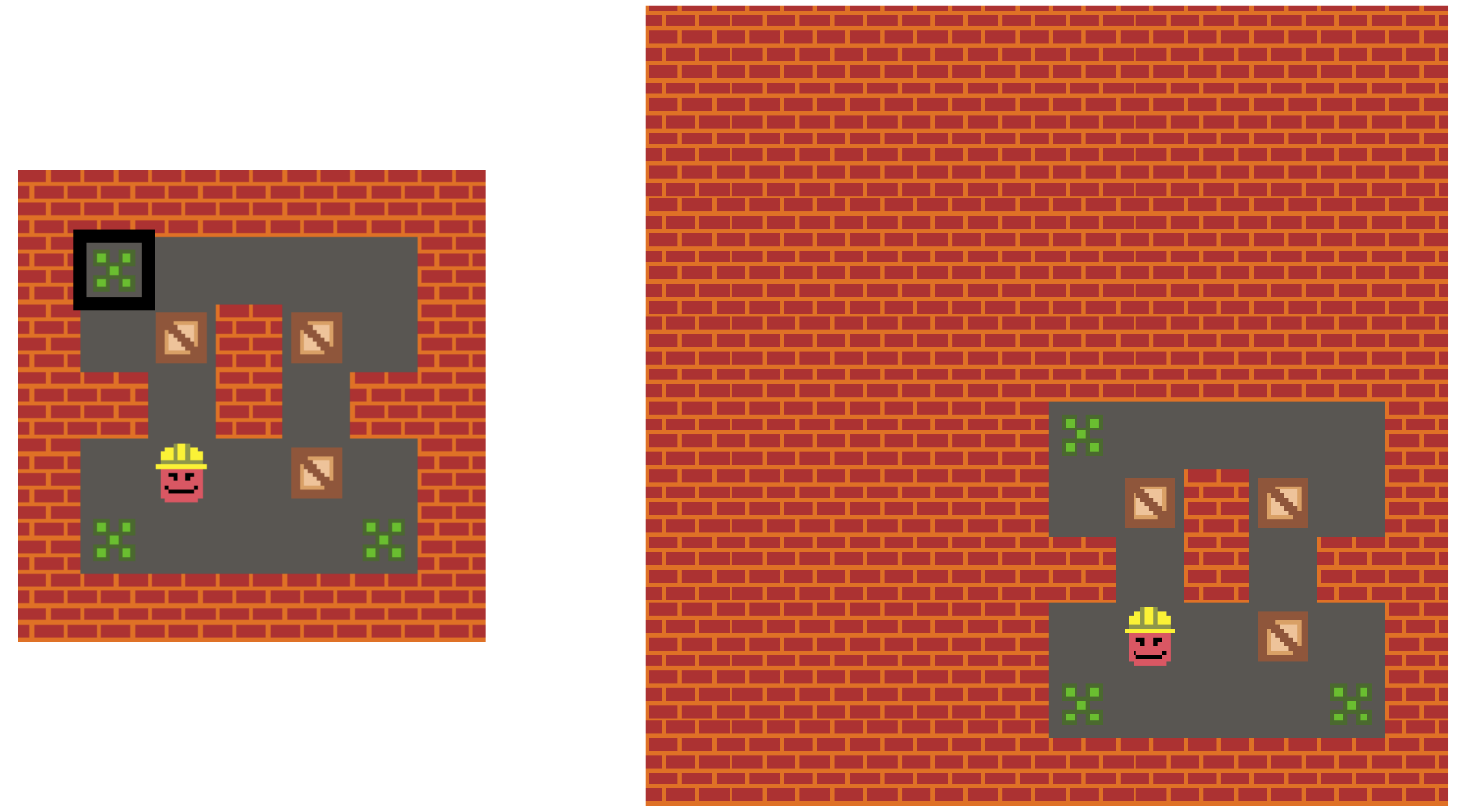}
    \caption{Location data is being transformed as an image translation}
    \label{fig:pcgrl_cropping}
\end{figure}

\begin{figure*}
    \centering
    \begin{subfigure}[t]{0.25\linewidth}
        \centering
        \includegraphics[width=\linewidth]{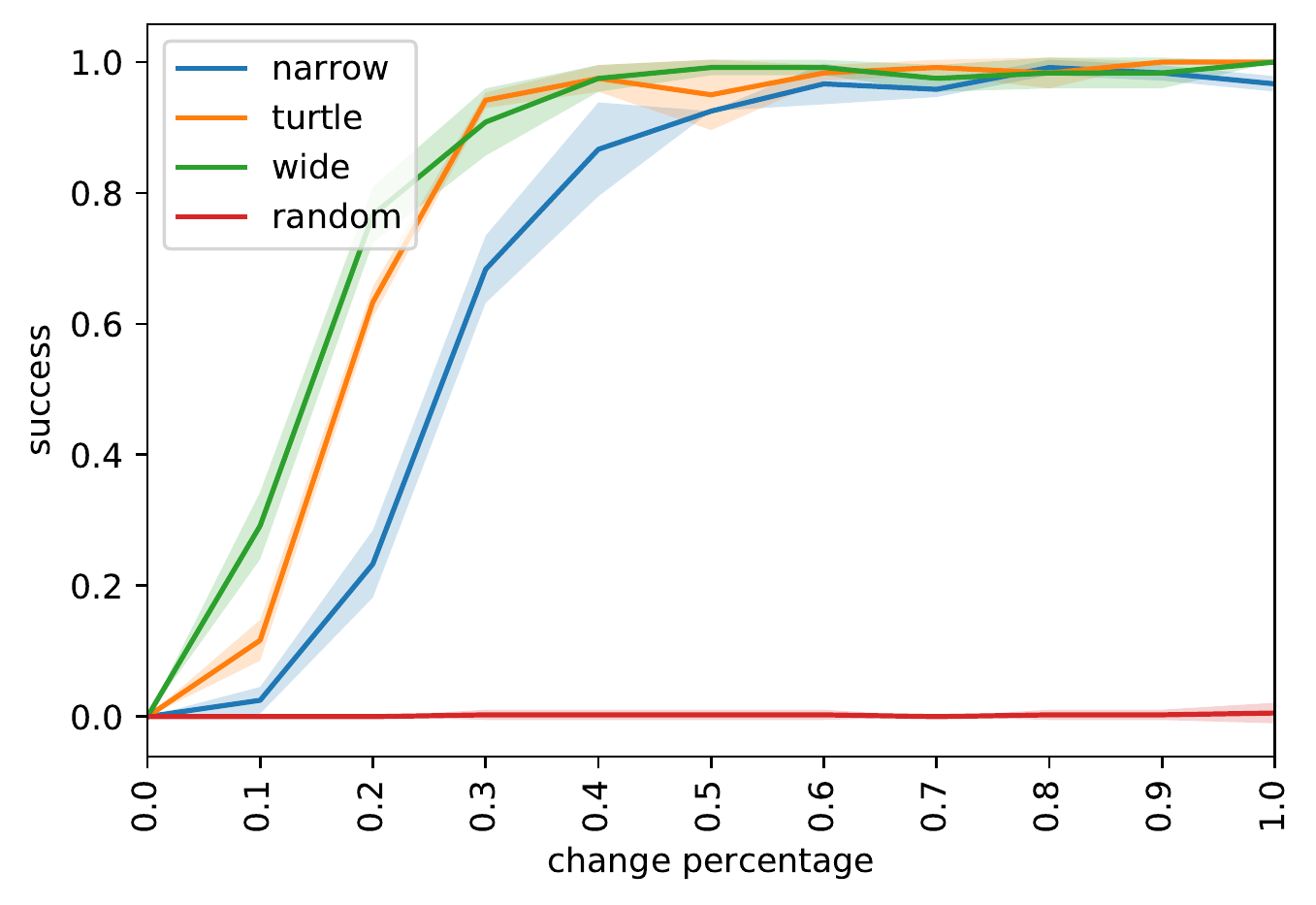}
        \caption{Binary}
        \label{fig:binary_sucess}
    \end{subfigure}
    \begin{subfigure}[t]{0.25\linewidth}
        \centering
        \includegraphics[width=\linewidth]{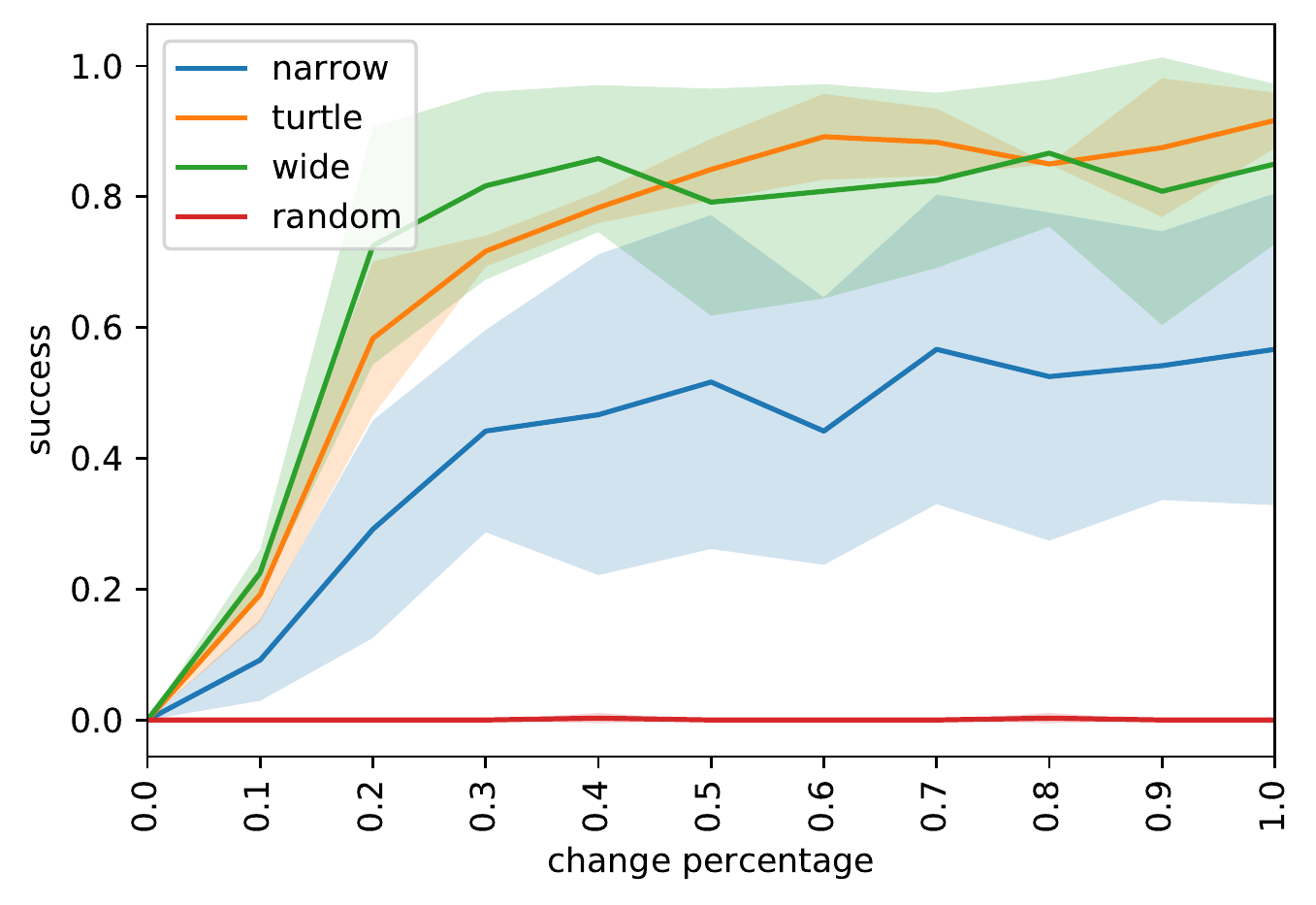}
        \caption{Zelda}
        \label{fig:zelda_success}
    \end{subfigure}
    \begin{subfigure}[t]{0.25\linewidth}
        \centering
        \includegraphics[width=\linewidth]{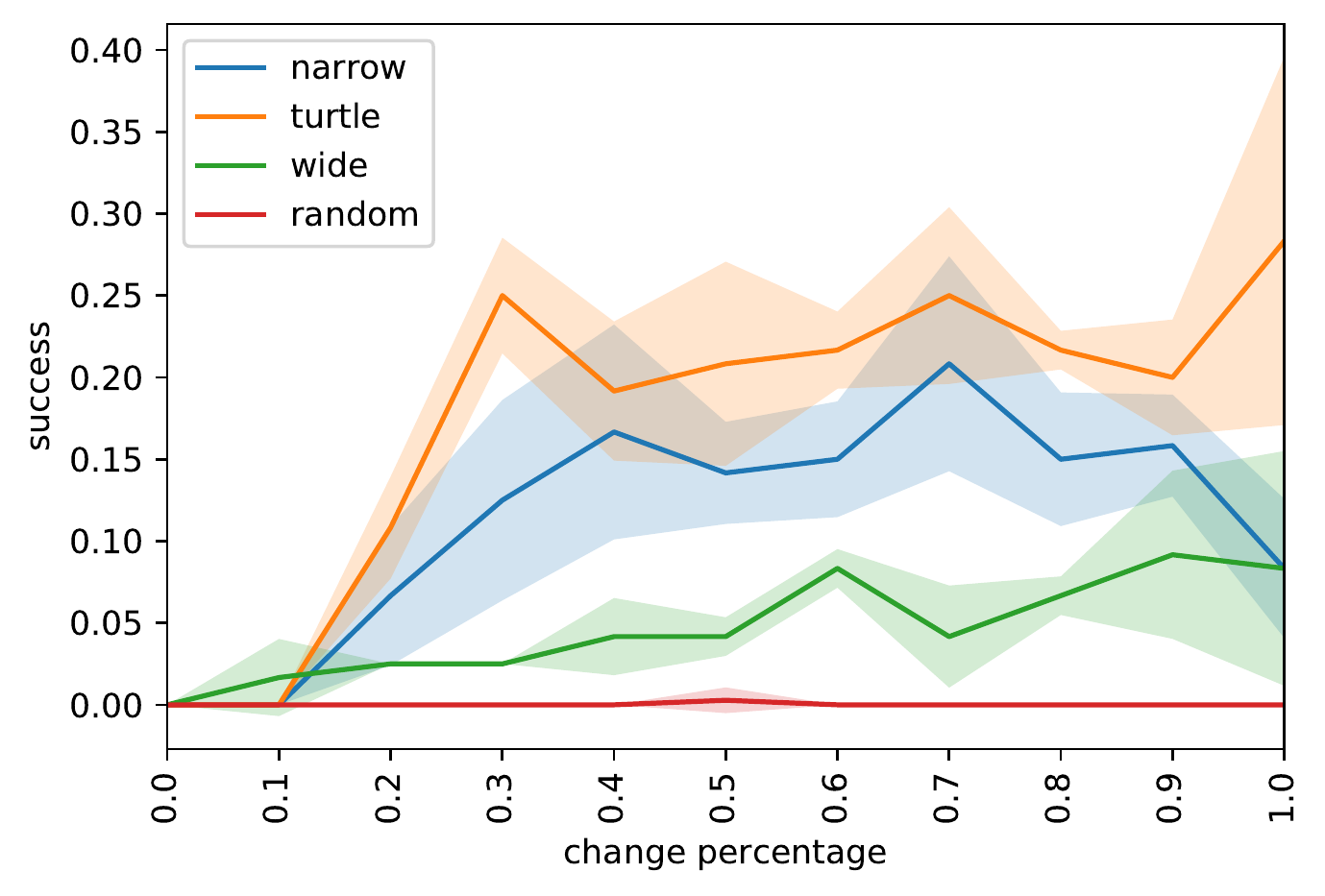}
        \caption{Sokoban}
        \label{fig:sokoban_success}
    \end{subfigure}
    \caption{The Success percentage of generating levels from random a initial state with respect to the change percentage.}
    \label{fig:success_graphs}
\end{figure*}

\subsection{Change Percentage}
The change percentage is an important parameter that defines how many tiles the agent is allowed to change as a percentage of the full size of the level. It limits the length of the episode during training so the agent cannot change all the tiles. We can think about it as similar to a discount factor: it defines how greedy the agent should be. For example: if the percentage is small, it means the agent is only allowed to make a very small amount of changes to the map, so the agent will end up learning more greedy actions to get higher short-term rewards. On the other hand, if the percentage is high, it means the agent is allowed to change as much of the map as possible, so the agent will end up learning a less greedy and more optimal solution to the problem. 

One might wonder why a more greedy agent is preferred over a more optimal agent. An agent that always makes the most optimal level, does not care about the initial random level layout. This means, that given complete information, the agent will converge to a single optimal solution. This is not the goal, as we desire an agent that acts as a designer that can transform an input level into a new level that is inspired by the input. There are multiple possible solutions to this; we decided to use the approach of limiting the agents actions to force it to make minimal changes to the environment as it is the most straight forward approach.


\section{Experiments}

Our PCGRL framework\footnote{\url{https://github.com/amidos2006/gym-pcgrl}} is implemented as an OpenAI Gym~\cite{brockman2016openai} interface, making it compatible with existing agents. We test the framework using three unique representations (Narrow, Turtle, and Wide) and problems (Binary, Zelda, and Sokoban). For all the problems, the reward function favors actions that help reach the goal of the problem while punishing those that move away from it. The problem terminates when the goal is reached. 
\begin{itemize}
    \item \textbf{Binary:} the easiest task, the goal is to modify a 2D map of solid and empty spaces such that the longest shortest path between any two points in the map increases by at least $X$ tiles (where $X=20$ in our experiments) and all the empty spaces are connected.
    \item \textbf{Zelda:} the goal is to modify a 2D level for Zelda, which is a port of the dungeon system in The Legend of Zelda (Nintendo, 1986) for the GVGAI framework~\cite{perez2019general}. The game's objective is to move the player to grab a key and then reach a door while avoiding getting killed by the moving enemies. The level must thus have exactly 1 player, 1 door, and 1 key (different game objects) and the player must be able to reach the key and then the door in at least $X$ steps (we set $X=16$). Enemies cannot spawn too close to the player (more than $3$ tiles away in our experiments). 
    \item \textbf{Sokoban:} the goal is to generate a 2D level for the Japanese puzzle game Sokoban (Thinking Rabbit, 1982), in which the player tries to push all crates toward certain target locations, while avoiding getting crates stuck against walls. The level must thus have exactly 1 player and the same number of crates and targets, all of them reachable by the player. This being the case, a Sokoban solver (a tree search algorithm (BFS and A*) with limited depth (around 5000 nodes)) is used to make sure levels can be solved in at least $X$ steps (we set $X=18$).
\end{itemize}

After some preliminary tests, we decided to fix the change percentage to 20\%, to encourage the agent to react, rather than to overwrite, the starting state. The starting state is randomly sampled from a random distribution which was tuned to fit each problem. For example: the player tile in Zelda and Sokoban have very low probability to appear in the starting state. Figures~\ref{fig:binary_start_examples},\ref{fig:zelda_start_examples}, and \ref{fig:sokoban_start_examples} show examples of different starting states for each of the three problems.

We randomize the starting position of the turtle and narrow representations to encourage generalization~\cite{justesen2018illuminating}. In narrow, each turn, the location is chosen at random so that the network will learn to react to any location. This ensures that all locations are visited with equal frequency during level-design. It need not be true during inference. We provide the agent with a one-hot encoding of the level map to help training.

\begin{table}
    \centering
    \resizebox{.55\linewidth}{!}{
        \begin{tabular}{|l||c|c|c|}
        \hline
                  & Narrow & Turtle & Wide\\
        \hline
        \hline
           Binary  & $15,918$ & $15,918$ & $246$\\
           \hline
           Zelda   & $8,451$ & $8,451$ & $256$\\
           \hline
           Sokoban & $585$ & $585$ & $214$\\
        \hline
        \end{tabular}
    }
    \caption{The sizes of the trained models in terms of thousands of parameters ($10^{3}$).}
    \label{tab:model_parameters}
\end{table}

For training, we use Proximal Policy Optimization (PPO)~\cite{schulman2017proximal} from \emph{Stable Baselines} (an improved implementation of OpenAI baselines~\cite{dhariwal2017openai}) to train our agents. We use two different architectures for the agent networks. The first is used for the binary and turtle representation. Its body consists of 3 convolution layers followed by a fully connected layer, and an additional fully connected layer for both the action and value heads similar to Google's ALE DeepQ architecture~\cite{mnih2015human}. The wide representation needed a different architecture due to its large action space (equal to the input space, e.g. 14x14 in binary). To solve that problem, we decided to use the similar architecture from \citeauthor{earle2019using}~\shortcite{earle2019using} work in playing SimCity. The body of the network consists of 8 convolutions, which is used directly for the action head, while the value head has 3 additional convolutions for Binary and Zelda and only 2 additional convolutions for Sokoban. Table~\ref{tab:model_parameters} shows the number of parameters for each representation and each game. For Narrow and Turtle, the network parameters are extremely big compared to wide, this is due to the presence of fully connected layer in the end which is not the case in the wide network. 

For each problem/representation experiment, we trained 3 different models to show the stability of the training. Each model was trained for 100 million frames. Since both Narrow and Turtle have location information, we encode this information as a translation of the map around that location which means the new map is twice as big as the old map and the center of that new map is the location needed. Figure~\ref{fig:pcgrl_cropping} shows the location information as a black rectangle on the left image. This information has been transformed as translation information in the right image where the position of that tile is the center of the new map.

\section{Results}

\begin{table}
    \centering
    \resizebox{.9\linewidth}{!}{
        \begin{tabular}{|l||c|c|c|}
        \hline
                  & Narrow & Turtle & Wide\\
        \hline
        \hline
           Binary & $29.7\% \pm 1.4\%$   & $19.7\% \pm 0.8\%$ & $15.6\% \pm 1\%$ \\
           \hline
           Zelda  & $39.25\% \pm 14.9\%$ & $21.6\% \pm 2.6\%$ & $21.8\% \pm 11.2\%$\\
           \hline
           Sokoban & $26.9\% \pm 1.6\%$  & $25.1\% \pm 2.2\%$ & $26.5\% \pm 2.3\%$\\
        \hline
        \end{tabular}
    }
    \caption{Percentage of map changed by the agent during inference when allowed unlimited changes.}
    \label{tab:change_percentages}
\end{table}

To analyze the trained models, we collect 40 generated levels for each model. To generate the levels, 40 different random level layouts are generated, the trained models are then tasked with modifying these random layouts into good levels. We analyze levels generated using different change percentages, ranging from $0\%$ to $100\%$, where the percentage represents the fraction of tiles the agent is allowed to change during inference. At each percentage, we analyze 120 different levels (40 generated levels per model).

Figure~\ref{fig:success_graphs} shows the percentage of generated levels that satisfy the goal criterion for each game. The horizontal axis represents the change percentage during inference while the vertical axis represents the percentage of successful levels. The solid line is the average of the three trained models while the shaded area is their standard deviation. We added a random agent to the mix. The random agent results were averaged between all three representation which turns out to be a 0\% success rate. We can see that some representations perform better than the others in some tasks while in others they perform the same. Predictably, when only allowed to make very few changes, the agent isn't able to design very good levels but possibly surprising is that the agent needs only to change at most $40\%$ of the tiles to design a successful level. This is especially interesting since the agent was only allowed to change $20\%$ of the tiles during training.

Figure~\ref{fig:zelda_success} shows that the narrow representation fails to get a high success rate compared to the other representations. We found that the narrow representation could design mostly good levels but fails to make their solution paths longer than $16$ steps. We believe that narrow might be struggling to improve the solution path due to the random nature of selecting the next tile during training and the need for long waits to make sure the system visits every single tile. We believe that it could be solved if the model were trained for a longer time (as it was still learning after 100 million frames).

\begin{figure}
    \centering
    \begin{subfigure}[t]{0.2\linewidth}
        \centering
        \includegraphics[width=.9\linewidth]{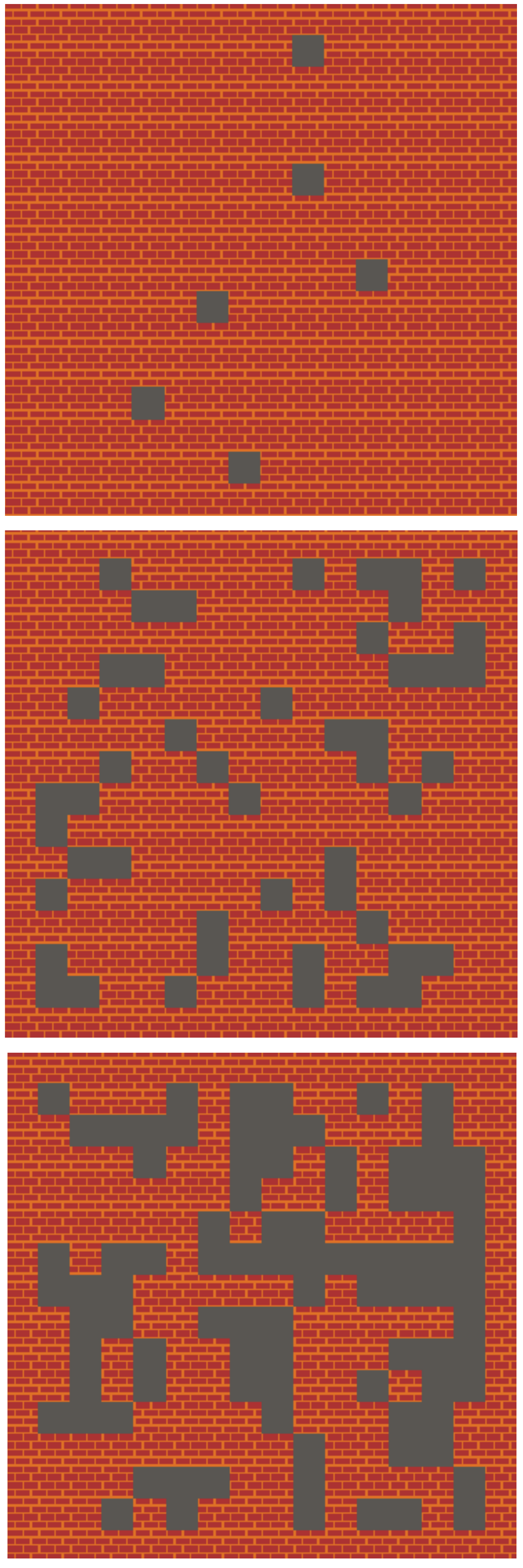}
        \caption{Initial}
        \label{fig:binary_start_examples}
    \end{subfigure}
    \begin{subfigure}[t]{0.2\linewidth}
        \centering
        \includegraphics[width=.9\linewidth]{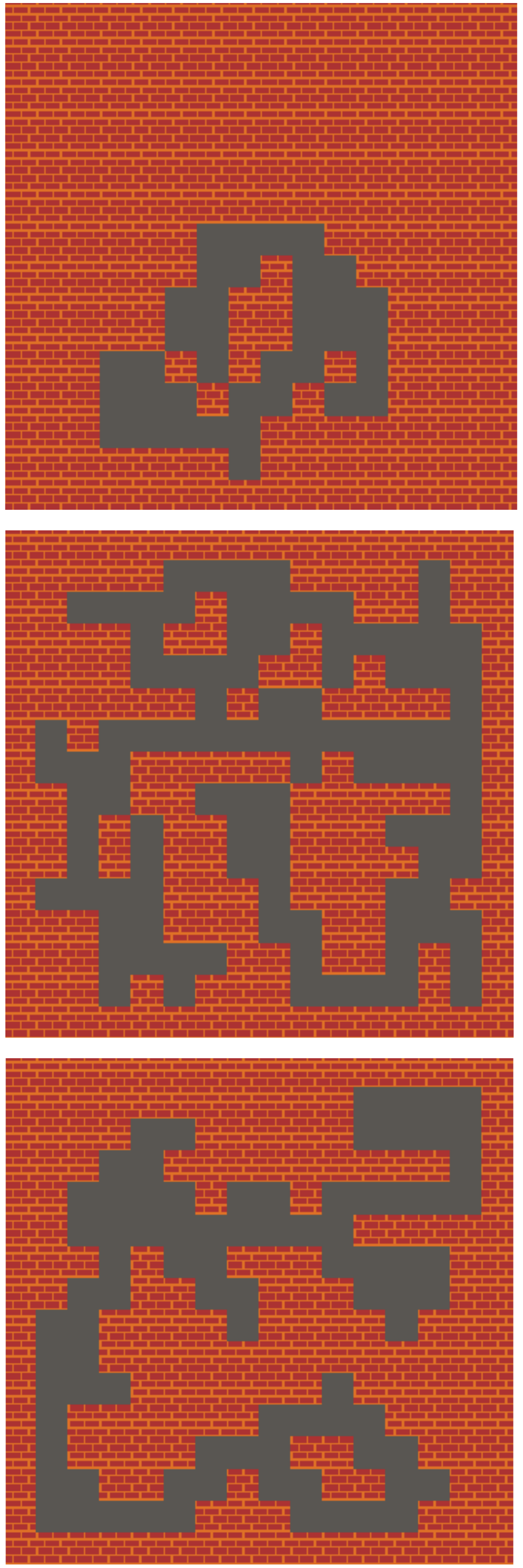}
        \caption{Narrow}
        \label{fig:binary_narrow_examples}
    \end{subfigure}
    \begin{subfigure}[t]{0.2\linewidth}
        \centering
        \includegraphics[width=.9\linewidth]{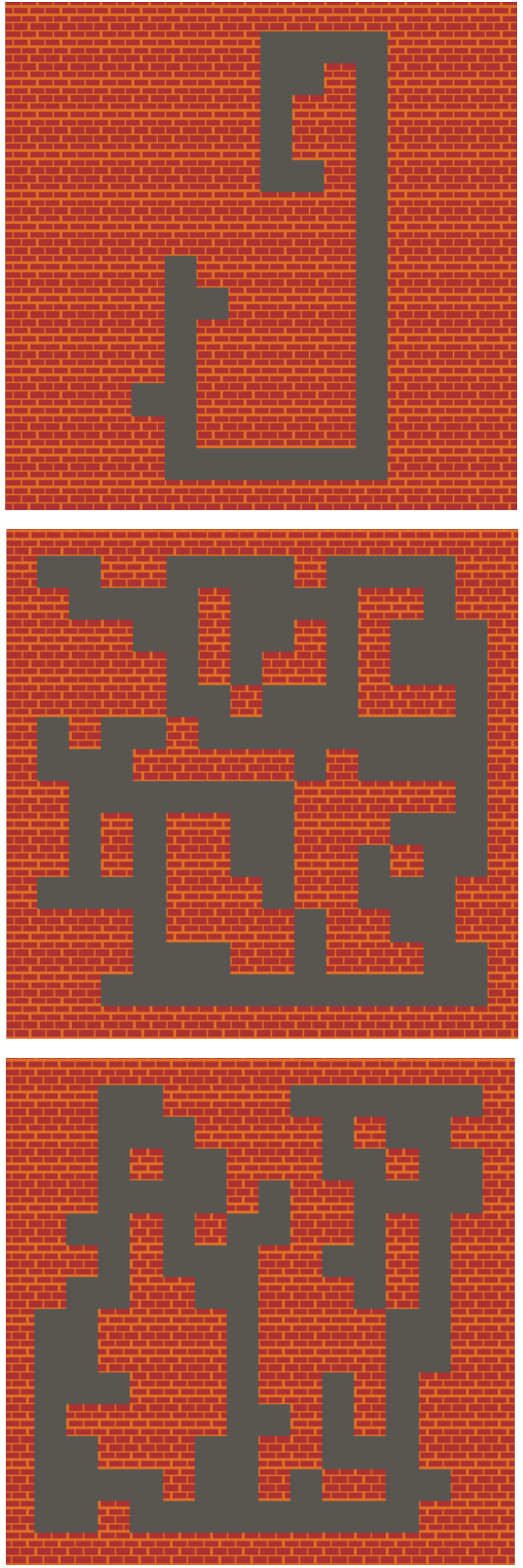}
        \caption{Turtle}
        \label{fig:binary_turtle_examples}
    \end{subfigure}
    \begin{subfigure}[t]{0.2\linewidth}
        \centering
        \includegraphics[width=.9\linewidth]{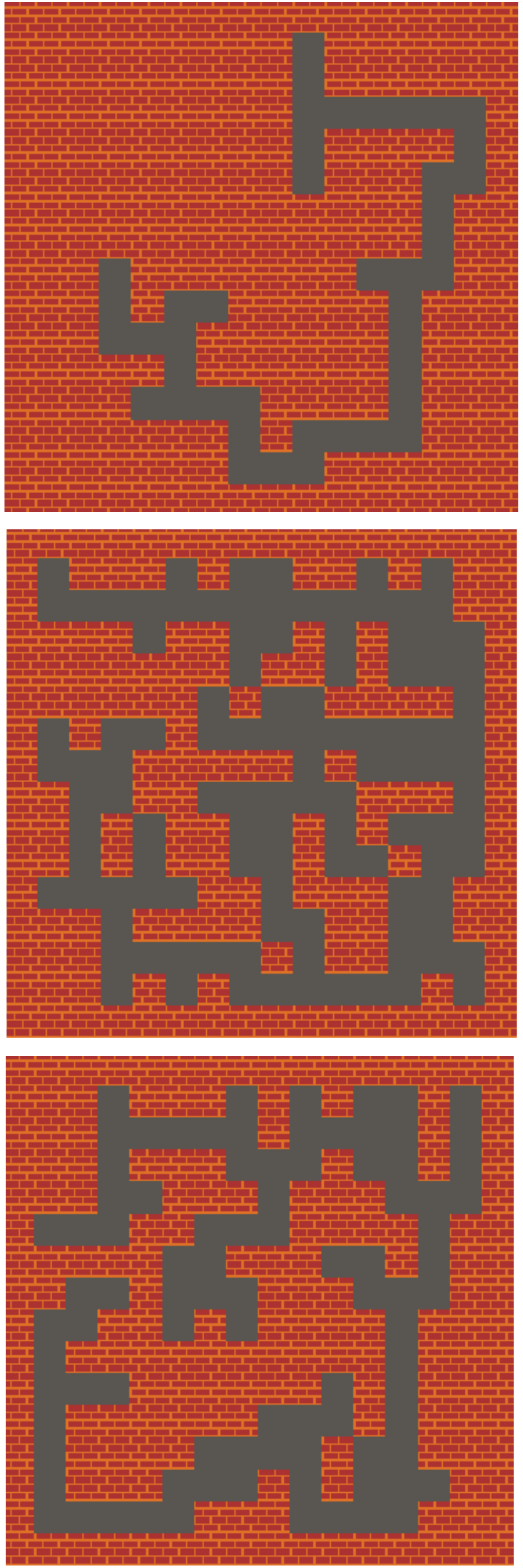}
        \caption{Wide}
        \label{fig:binary_wide_examples}
    \end{subfigure}
    \caption{Binary generated examples using different representation for the same starting state.}
    \label{fig:binary_examples}
\end{figure}

Figure~\ref{fig:sokoban_success} shows the result of generating Sokoban levels. We see that the success rate is relatively low, especially in the wide representation. We looked into the generated levels and found that most of them are easy levels that can be solved in few steps (less than $18$). Narrow models generate 86.7\% solvable 
, Turtle models generate 88.3\% solvable 
, while Wide models generate 67.5\% solvable levels 
at 100\% change percentage. Based on this, we think that this problem could be solved if the model were trained longer and rewarded based on a more powerful Sokoban solver. The current solver only solves simple levels to be quick. 

We suspect that the performance drop that appears when using the wide representation on Sokoban (as compared to Zelda and Binary) might be due to the relative shortage of parameters in the model's critic head. Due to the small map-size of our Sokoban levels (compared to Binary and Zelda), and since the full network is convolutional with no fully connected layers, our model needs fewer (strided convolutional) layers in the critic head to arrive at a scalar value estimate than it does in Binary and Zelda. We believe that supplementing the model’s critic head with some additional non-strided convolutional layers, or adding parameters via increased number of channels, could make this model competitive compared to the others.

We looked at how many changes the trained models are making on average. Table~\ref{tab:change_percentages} shows the percentage of changed tiles in the map if the agent has no upper bound (change percentage is equal to 100\% during inference). We can see that most of the agents still don't make a lot of changes, which proves that the models are reacting to the input map instead of overwriting it with an optimal solution. 

\begin{figure}
    \centering
    \begin{subfigure}[t]{0.2\linewidth}
        \centering
        \includegraphics[width=\linewidth]{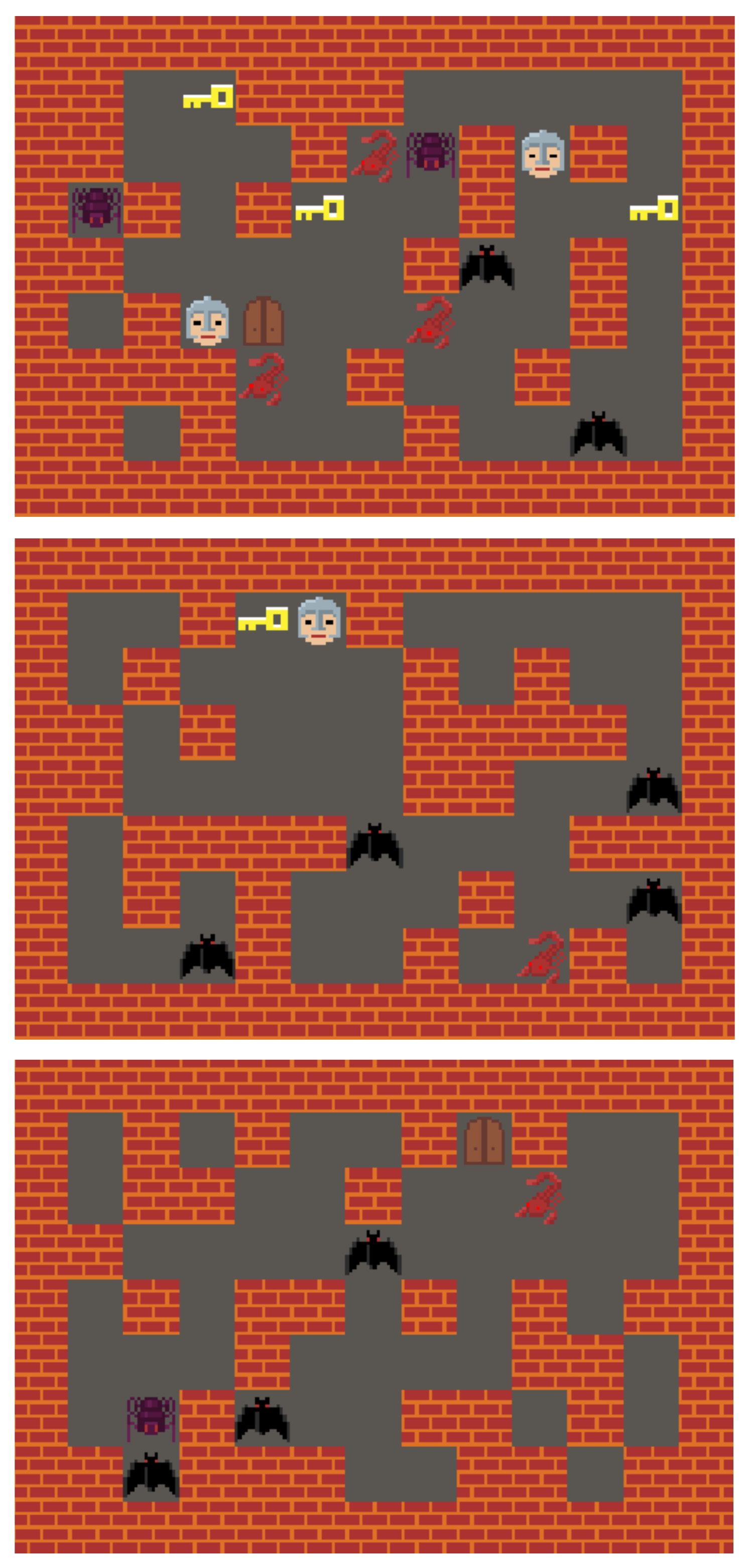}
        \caption{Initial}
        \label{fig:zelda_start_examples}
    \end{subfigure}
    \begin{subfigure}[t]{0.2\linewidth}
        \centering
        \includegraphics[width=\linewidth]{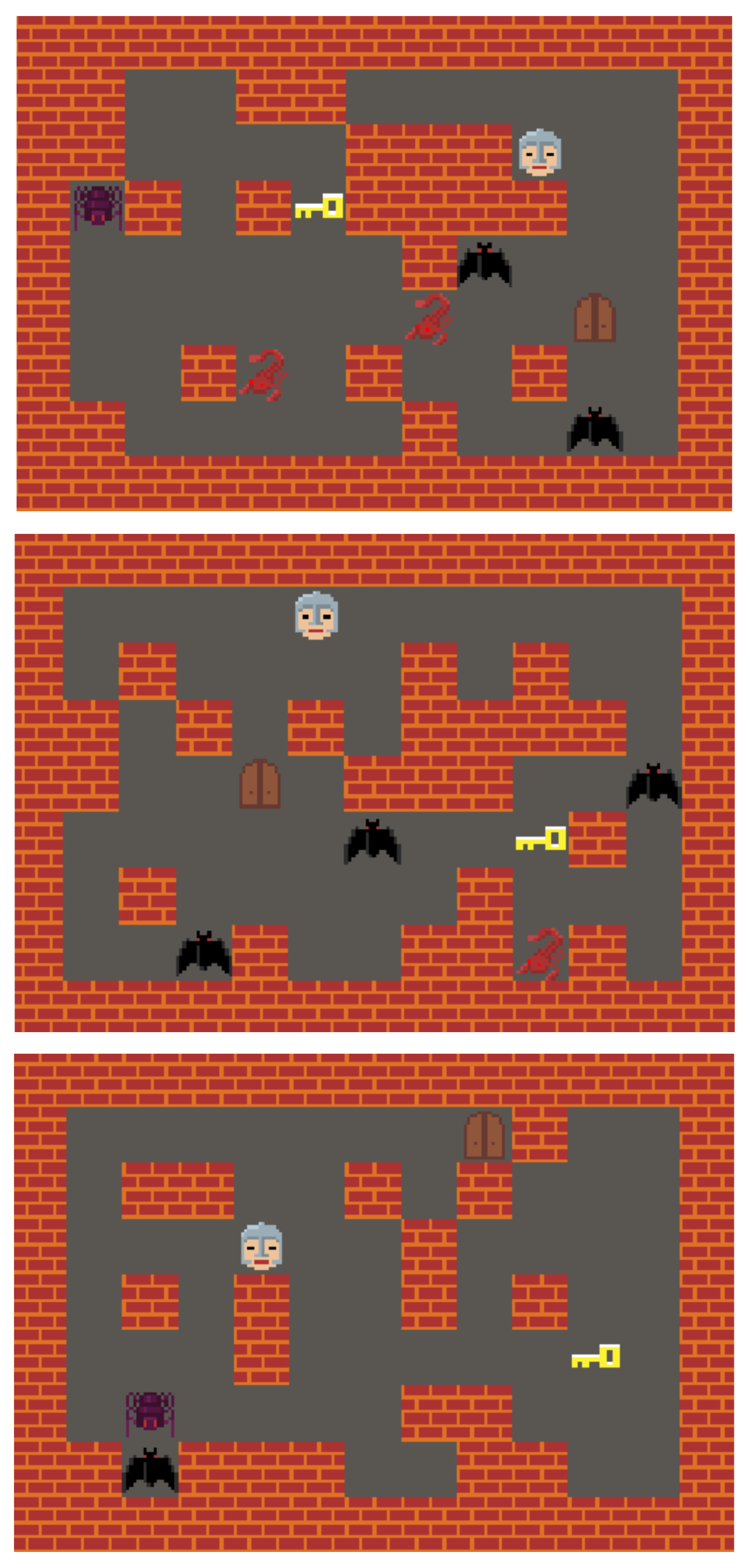}
        \caption{Narrow}
        \label{fig:zelda_narrow_examples}
    \end{subfigure}
    \begin{subfigure}[t]{0.2\linewidth}
        \centering
        \includegraphics[width=\linewidth]{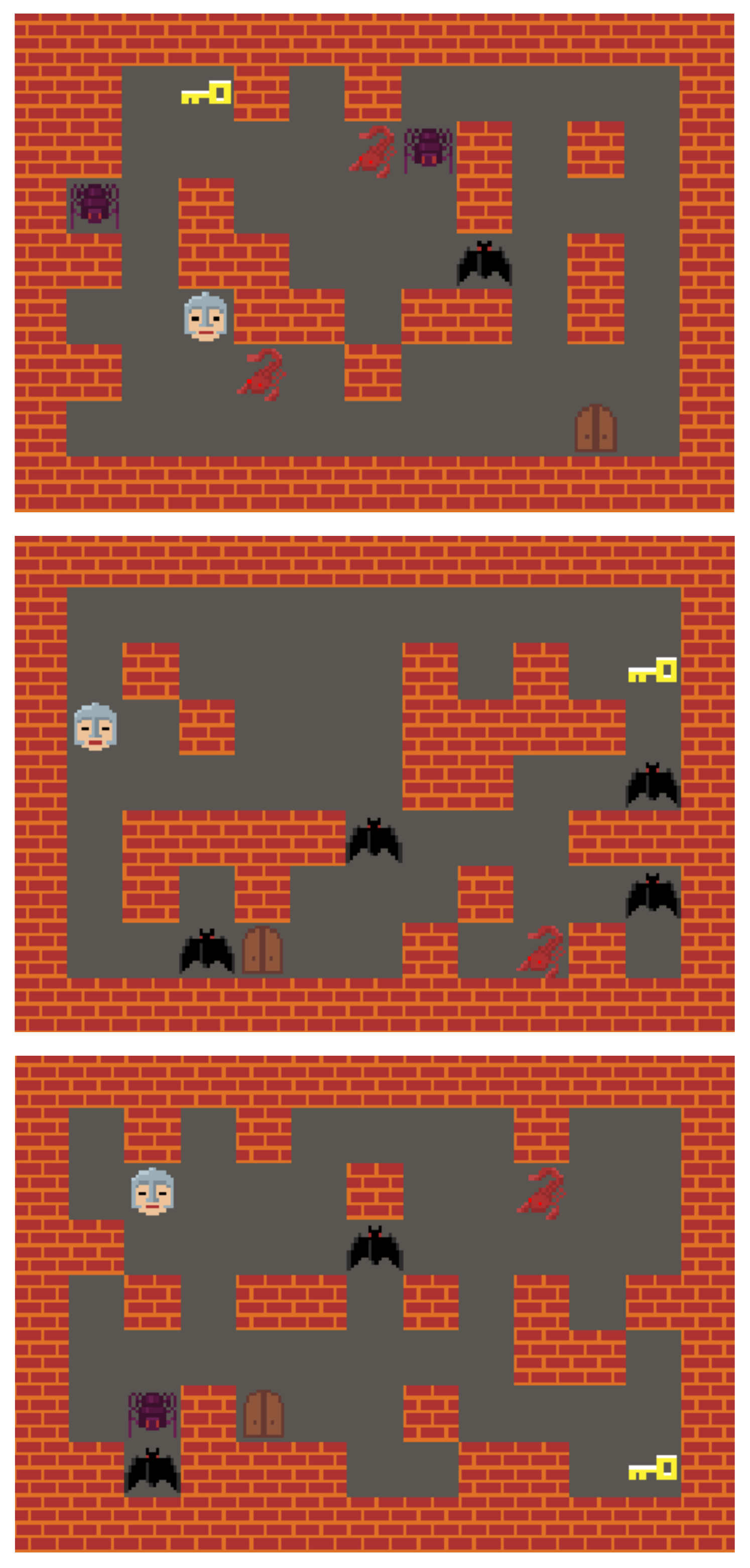}
        \caption{Turtle}
        \label{fig:zelda_turtle_examples}
    \end{subfigure}
    \begin{subfigure}[t]{0.2\linewidth}
        \centering
        \includegraphics[width=\linewidth]{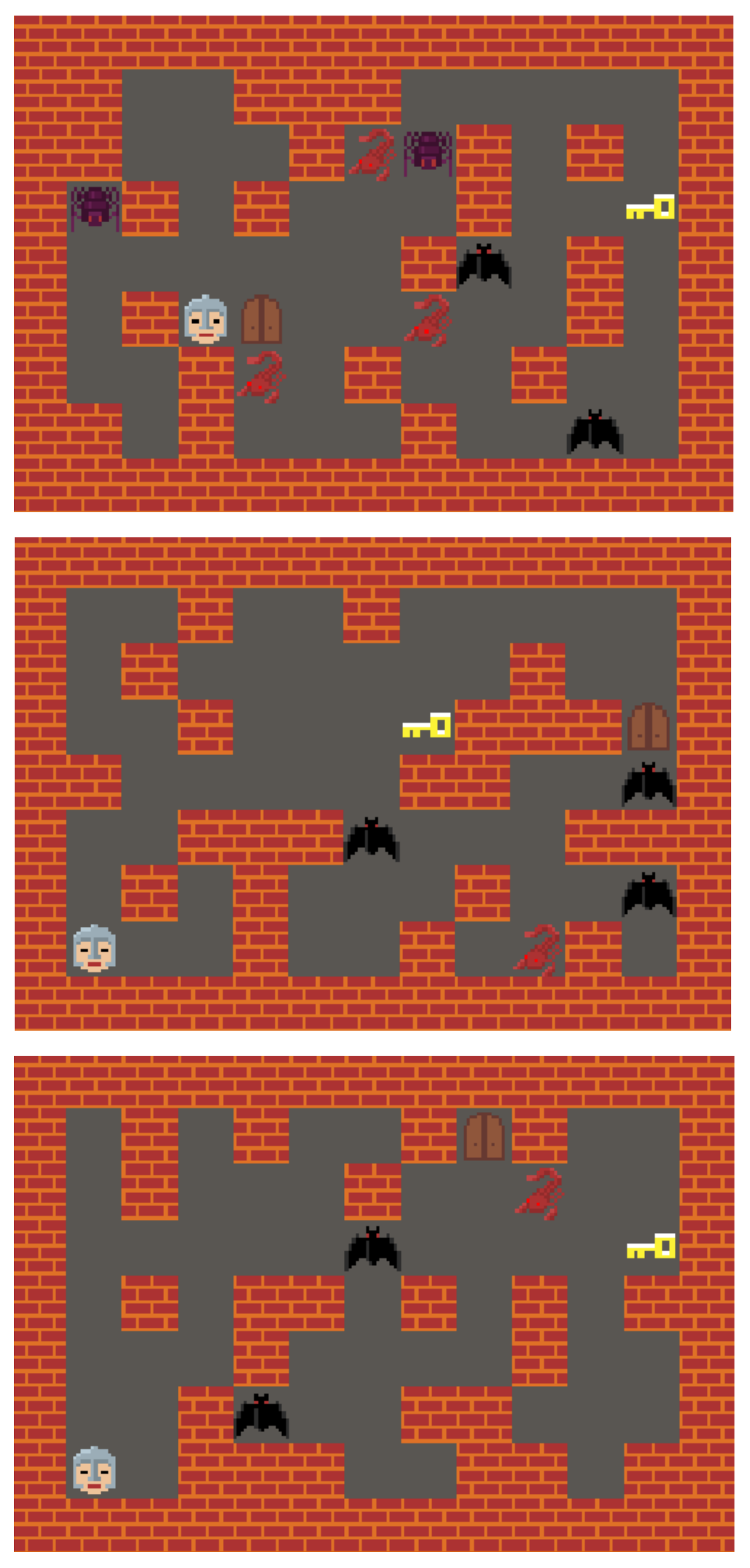}
        \caption{Wide}
        \label{fig:zelda_wide_examples}
    \end{subfigure}
    \caption{Zelda generated examples using different representation for the same starting state.}
    \label{fig:zelda_examples}
\end{figure}

\begin{figure}
    \centering
    \begin{subfigure}[t]{0.2\linewidth}
        \centering
        \includegraphics[width=.8\linewidth]{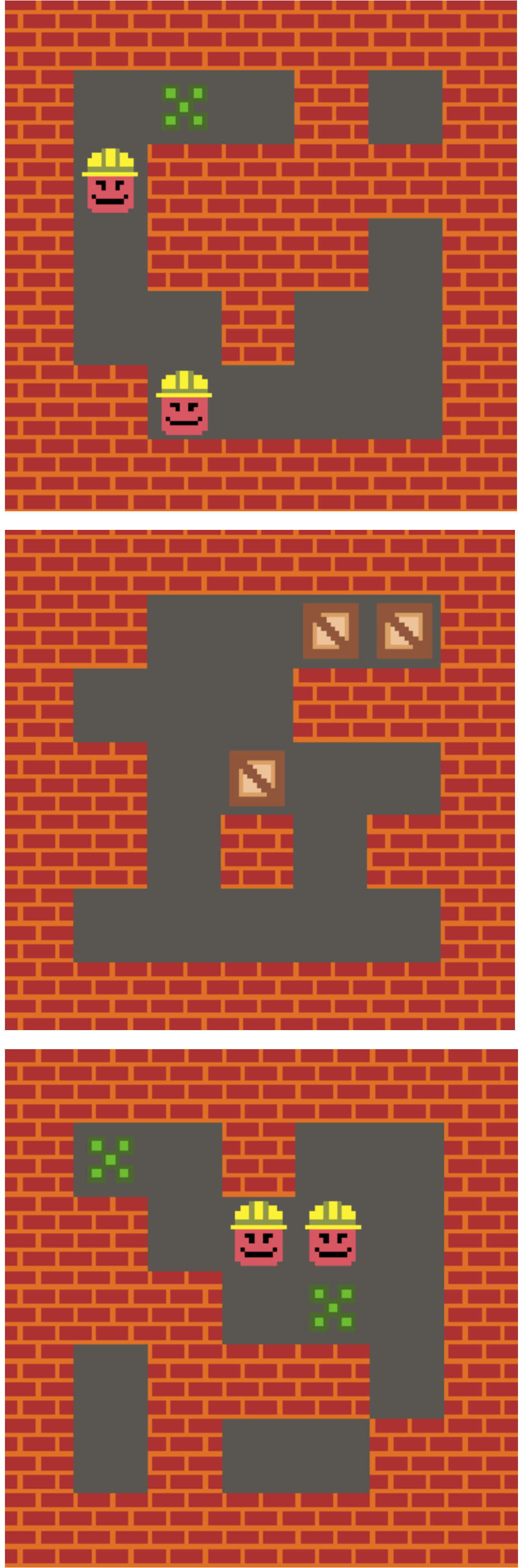}
        \caption{Initial}
        \label{fig:sokoban_start_examples}
    \end{subfigure}
    \begin{subfigure}[t]{0.2\linewidth}
        \centering
        \includegraphics[width=.8\linewidth]{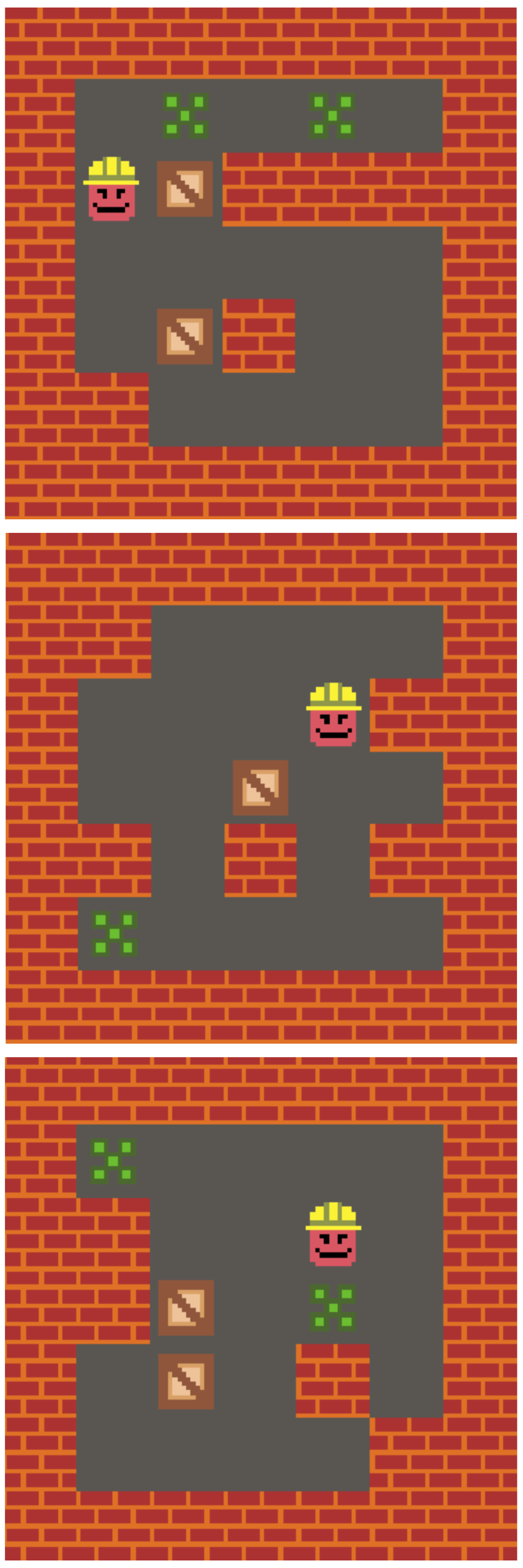}
        \caption{Narrow}
        \label{fig:sokoban_narrow_examples}
    \end{subfigure}
    \begin{subfigure}[t]{0.2\linewidth}
        \centering
        \includegraphics[width=.8\linewidth]{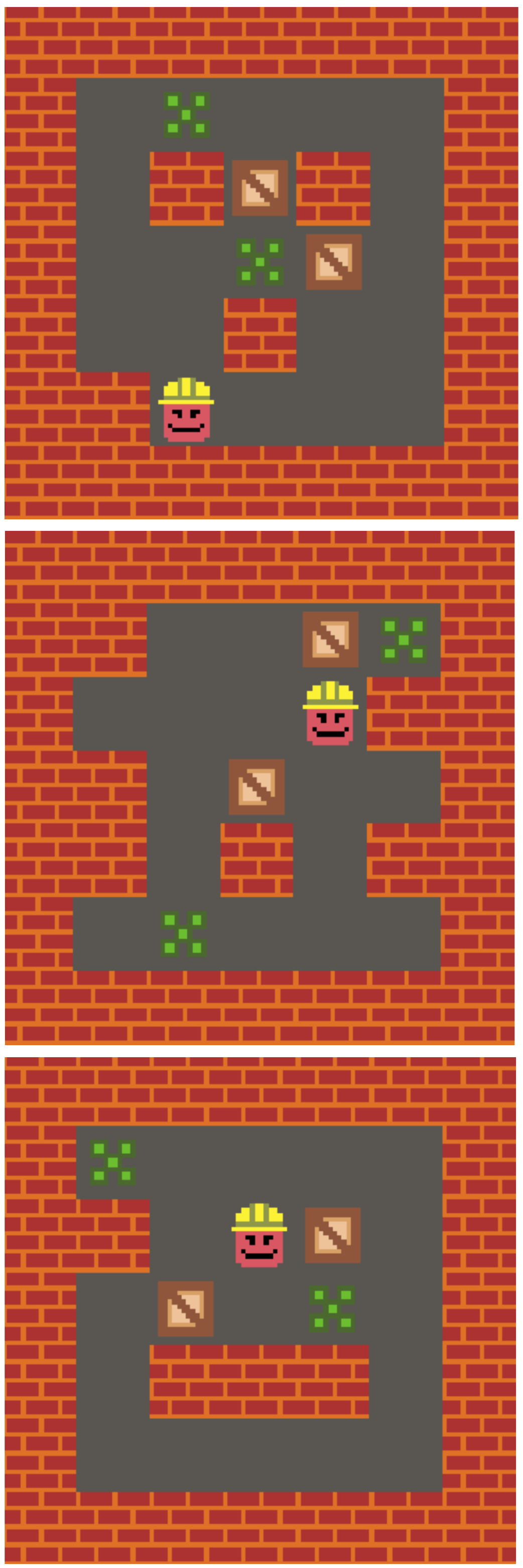}
        \caption{Turtle}
        \label{fig:sokoban_turtle_examples}
    \end{subfigure}
    \begin{subfigure}[t]{0.2\linewidth}
        \centering
        \includegraphics[width=.8\linewidth]{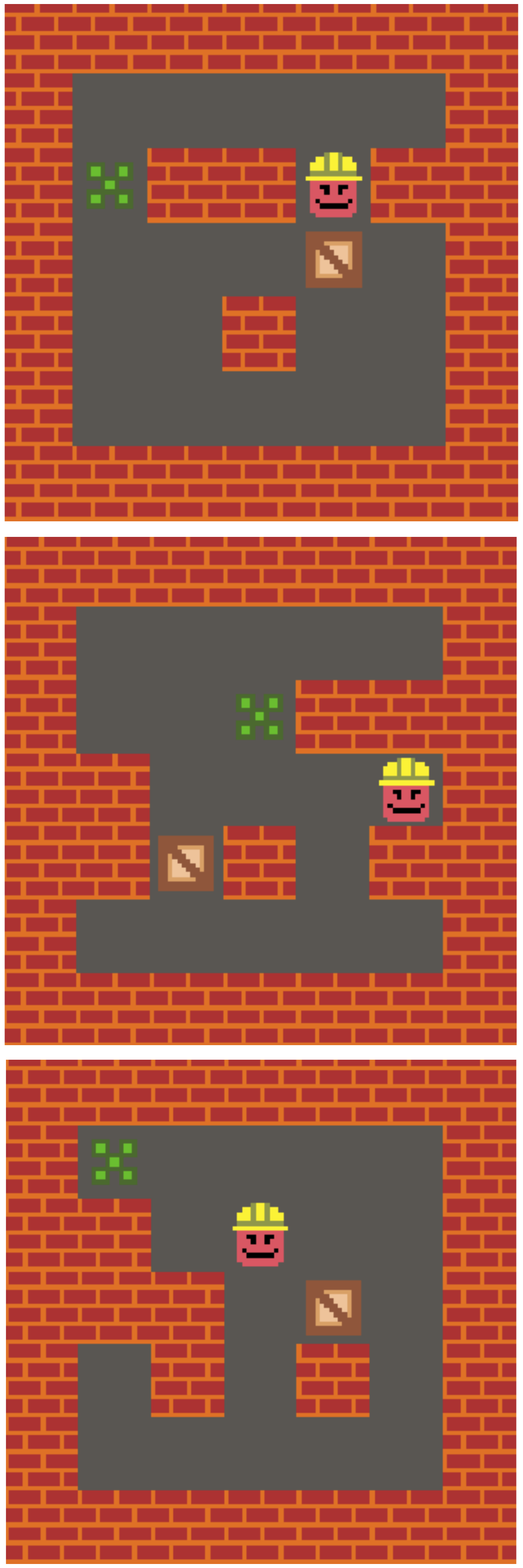}
        \caption{Wide}
        \label{fig:sokoban_wide_examples}
    \end{subfigure}
    \caption{Sokoban generated examples using different representation for the same starting state.}
    \label{fig:sokoban_examples}
\end{figure}

Figures~\ref{fig:binary_examples},\ref{fig:zelda_examples}, and \ref{fig:sokoban_examples} show the results from running a trained model on the Binary, Zelda, and Sokoban problems respectively. In these experiments, we fix the starting state and run the trained model for each representation. We run the model with 100\% change percentage instead of the 20\% change percentage (used during training). The reason for the increase in the change percentage is to allow the algorithm to make more changes in case it started from very bad state.

As shown in figure~\ref{fig:success_graphs}, different representations do not seem to have a large effect on success across every problem. The real difference can be seen in design styles. This is especially clear in the Binary problem (figure~\ref{fig:binary_examples}). In the Binary problem, all representations achieve similar success percentages, but each results in a stylistically distinct set of generated levels. In the narrow representation, the agent has no control over when it is allowed to make a change in a specific area. From this it seems to have learned to make more frequent sub-optimal changes rather than waiting for the ideal change. This makes the narrow representation the least ``controlled'' by the initial conditions and to have the most varied results. The turtle representation has control over where it goes, but only loosely, as navigation is difficult and it's possible for the game to end if it does not move efficiently. It seems to favor updating the board in a spiral manner. Lastly, the wide representation is the most efficient with its changes, with its designs being closest to the original pattern. This agent could always control a change exactly, so it never needed to risk ending the game with too many changes. The downside of using the wide representation is that it must learn a much larger action space than the others.

\section{Discussion}

The two major conceptual differences between this work and almost all published research on PCG is to search in content generator space rather than content space, and to see content generation as an iterative improvement problem. Searching in content generator space was prefigured by the Procedural Procedural Level Generator Generator (PPLGG), which evolves Super Mario Bros level generators~\cite{kerssemakers2012procedural}. However, the search in PPLGG proceeds through interactive evolution rather than RL, and the generators are simple multi-agent systems. The paradigm presented here is more scalable, not least for its automatic learning. 

We only discussed one type of content(game levels), as a case study, to show that the technique works and to demonstrate how one may go about transforming a specific (and in this case, canonical) PCG problem into an RL problem. If one has a PCG problem definition and reward function, one can readily adopt our narrow, turtle, or wide agent–environment interfaces, and run the PCGRL algorithm on any content generation problem.

It is particularly exciting to see the multiple uses PCGRL could have in mixed-initiative editing tools. As the interface of a trained model is to make a single small change intended to improve the level, this fits in very well with a turn-taking paradigm. One could imagine multiple trained PCGRL agents working as suggestion engines, pointing out where to improve the current design or as ``brushes,'' applied by the human user to e.g. increase difficulty in a region.

One of the main reasons that PCG approaches based on search/optimization have not been applied widely in the game industry is that they take too much time. And it is indeed true that if any kind of simulation is incorporated into an evaluation function, the search for good game content can be costly. Here, PCGRL offers the possibility of moving most of the time consumption from inference to training, or, in game development terms, from runtime to development. This might make PCGRL a viable choice for runtime content generation in games. 
The problem of designing an appropriate reward scheme remains, however, and is similar to that of designing an evaluation function in search-based PCG. 

\section{Conclusion}
We propose a framework for generating content using reinforcement learning. We test the idea on level generation for three different problems (Binary, Zelda, and Sokoban). To make the system work, we adapt three different representations (Narrow, Turtle, and Wide) from~\citeauthor{bhaumik2019tree}'s~\shortcite{bhaumik2019tree} work. These representations transform the generation process into a MDP which is easy to implement as an OpenAI Gym interface~\cite{brockman2016openai}. We train three models using the PPO algorithm for each problem/representation. For the Binary problem, all models performed pretty well given enough changes, and the main difference was the style of the generated content. For Zelda and Sokoban, the agent struggled to design hard levels but was able to generate a high amount of playable levels. This problem could be tackled by using more powerful game solvers and training the models for longer, as the models were still learning after 100 million frames.

This work introduces the main concept and principle of a new approach for generating content using reinforcement learning. It opens the door to the application of many fruitful ideas from RL to the world of PCG, such as self-play agents (where the same agent plays against itself in a competitive game to improve the content), collaborative agents (where two or more agents work together to generate the level), hierarchical agents (where the agent has separate modules for different functionalities; for example one for movement and another for modification), etc. In future work, we would like to extend the framework to support mixed-initiative approaches by preventing the agent from overwriting human content, and thus allowing for the same kind of human-AI collaboration as in \citeauthor{guzdial2018co}'s~\shortcite{guzdial2018co} work.

\section*{Acknowledgments}
Ahmed Khalifa acknowledges the financial support from NSF grant (Award number 1717324 - ``RI: Small: General Intelligence through Algorithm Invention and Selection.'').

\bibliography{AAAIbiblo}
\bibliographystyle{aaai}

\end{document}